\documentclass[conference]{IEEEtran}
\ifCLASSINFOpdf
\else
\fi

\usepackage{color}
\usepackage{algorithm,algorithmicx}
\usepackage{algpseudocode,empheq,hhline}
\usepackage{amsmath,amsthm,amssymb,url,graphicx,rotating,ifthen,epsfig,array,caption,color}
\usepackage[bookmarks=false]{hyperref}
\usepackage{todonotes}
\usepackage{graphicx}
\usepackage{caption}
\usepackage{subcaption}
\usepackage{balance}
\usepackage{multirow}
\usepackage{enumitem}

\makeatletter
\newcommand{\putindeepbox}[2][0.7\baselineskip]{{%
    \setbox0=\hbox{#2}%
    \setbox0=\vbox{\noindent\hsize=\wd0\unhbox0}
    \@tempdima=\dp0
    \advance\@tempdima by \ht0
    \advance\@tempdima by -#1\relax
    \dp0=\@tempdima
    \ht0=#1\relax
    \box0
}}
\makeatother

\newcounter{eqn}

\def\N{{\mathbb N}}

\def\P{{\mathbb P}}
\def\G{{\cal N}}

\def\R{{\mathbb R}}
 
\def\x{\mathbf{x}}

\newcommand{\et}[2]{\mathbb{T}_{({#2}\rightarrow{#1})}}

\begin{document}
%
\title{Optimal resampling for the noisy OneMax problem}




\author{
\IEEEauthorblockN{Jialin Liu}
\IEEEauthorblockA{University of Essex\\
Colchester CO4 3SQ\\
United Kingdom\\
\url{jialin.liu@essex.ac.uk}}
\and
\IEEEauthorblockN{Michael Fairbank}
\IEEEauthorblockA{University of Essex\\
Colchester CO4 3SQ\\
United Kingdom\\
\url{m.fairbank@essex.ac.uk}}
\and
\IEEEauthorblockN{Diego P\'erez-Li\'ebana}
\IEEEauthorblockA{University of Essex\\
Colchester CO4 3SQ\\
United Kingdom\\
\url{dperez@essex.ac.uk}}
\and
\IEEEauthorblockN{Simon M. Lucas}
\IEEEauthorblockA{University of Essex\\
Colchester CO4 3SQ\\
United Kingdom\\
\url{sml@essex.ac.uk}}
}


%


\maketitle

\begin{abstract}
The OneMax problem is a standard benchmark optimisation problem for a binary search space. Recent work on applying a Bandit-Based Random Mutation Hill-Climbing algorithm to the noisy OneMax problem 
showed that it is important to choose a good value for the resampling number to make a careful trade off between taking more samples in order to reduce noise, and taking fewer samples to reduce the total computational cost. 
This paper extends that observation 
by deriving an analytical expression for the running time of the Random Mutation Hill-Climbing algorithm with resampling applied to the noisy OneMax problem, and showing both theoretically and empirically that the optimal resampling number increases with the number of dimensions in the search space.

\end{abstract}

\begin{IEEEkeywords}
Noisy OneMax, resampling, Random Mutation Hill-Climber (RMHC)
\end{IEEEkeywords}

%


\section{Introduction}

Evolutionary Algorithms (EA) have been widely used in both continuous and discrete domains~\cite{muhlenbein1992genetic,droste2002analysis,arnold2006general}.

Resampling\footnote{In this paper, ``resampling'' refers to the multiple re-evalutions of a solution.} has proved to be a powerful tool in improving
the local performance of EAs in noisy optimisation 
~\cite{arnold2006general,beyer2013theory,astete2016simple},
and different resampling rules have been applied to a variety of EAs in continuous noisy optimisation, as studied in \cite{astete2013log,liu2015portfolio}.
Akimoto et al.~\cite{akimoto2015analysis} concluded that the running time for an adapted algorithm using resampling to solve a problem with additive Gaussian noise is similar to the runtime in the noise-free case when multiplying by
a factor $\log{n}$, where $n$ is the problem dimension.

Previous work on solving the OneMax problem~\cite{schaffer1991crossover}
has concentrated on using a (1+1)-Evolution Algorithm (EA)~\cite{droste2004analysis,qian2014effectiveness}. The OneMax problem 
with \emph{One-bit} noise (exactly one uniformly selected bit changes with probability $p \in (0,1)$ due to the noise) has been studied previously by Droste~\cite{droste2004analysis}.
Qian et al.~\cite{qian2014effectiveness} claimed that resampling wasn't beneficial in optimising OneMax with additive Gaussian noise using (1+1)-EA.
Recently, Liu et al.~\cite{liu2016bandit,liu2017bandit} applied a bandit-based RMHC to the noisy 
OneMax problem
, and showed that it was important to choose an optimal resampling number, so as to compromise 
the reduction in sampling noise against the cost of doing so.

The main contribution of this work is the analysis of the optimal resampling number in the OneMax problem, 
in the presence of additive Gaussian noise. We show that the optimal resampling number increases with the problem dimension.

The paper is structured as follows.
Section \ref{sec:bg} provides a brief review of the related work and describes our noisy OneMax problem.
Section \ref{sec:rmhcr} explains the modified Random Mutation Hill-Climbing algorithm used in the noisy context.
Section \ref{sec:analysis} analyses the optimal resampling number in the defined noisy OneMax problem.
Experimental results are presented and discussed in Section \ref{sec:xp}.
Finally, Section \ref{sec:conc} concludes the work.

\section{Background}\label{sec:bg}
This section is organised as follows. Section \ref{sec:rmhc} presents the original Random Mutation Hill-Climbing algorithm and its relation to (1+1)-EA. Section \ref{sec:onemax} recalls the OneMax problem, then a noisy variant of OneMax problem is defined in Section \ref{sec:nonemax}. More related literatures in solving different noisy variants of OneMax problems are discussed in Section \ref{sec:rw}

\subsection{Random Mutation Hill-Climbing}\label{sec:rmhc}

The Random Mutation Hill-Climbing (RMHC), also called Stochastic Hill Climbing, is a derivative-free optimisation method mostly used 
in discrete domains~\cite{lucas2003learning,lucas2005learning}.  
RMHC can also be seen as an evolutionary algorithm in which there is a population of just one individual, and at each generation a child genome is formed from the current (best-so-far) individual by mutating exactly one gene, chosen uniformly at random.  
After mutation, the mutated child genome replaces its parent if its fitness value is improved or equivalent. In other words, RMHC randomly selects a neighbour candidate in the search space (where \emph{neighbour} means it differs in exactly one gene) and updates its current candidate using a fitness-comparison-based 
method.

Borisovsky and Eremeev \cite{borisovsky2008comparing} proved that under some conditions
, RMHC outperforms other EAs in terms of the probability of finding an \emph{anytime} solution on several problems including OneMax.

(1+1)-EA 
is a variant of RMHC, the only difference being that \emph{every} gene of the current individual's genome mutates with a certain probability at each generation.\footnote{However note that some communities refer to RMHC as ``(1+1)-Evolutionary Algorithm (EA)'', but we will avoid that description.} (1+1)-EA has been widely used in both discrete and continuous optimisation problems \cite{borisovsky2008comparing}. A variant of (1+1)-EA is the closely related ``(1+1)-Evolutionary Strategy (ES)''~\cite{Beyer1995Toward}, 
which uses a self-adaptive mutation step-size and is applicable to real-valued search spaces.

Despite its simplicity, RMHC often competes surprisingly well with 
more complex algorithms~\cite{mitchell1993will}, especially when deployed with random restarts.
For instance, Lucas and Reynolds evolved Deterministic Finite Automata (DFA)~\cite{lucas2003learning,lucas2005learning} 
using a multi-start RMHC algorithm with very competitive results, outperforming more complex evolutionary algorithms, and for some classes of problems also outperforming the state of the art Evidence-Driven State Merging (EDSM)~\cite{cicchello2002beyond} algorithms. 

\subsection{OneMax problem}\label{sec:onemax}
The OneMax problem~\cite{schaffer1991crossover} is a standard benchmark optimisation problem for a binary search space and has been deeply studied in the previous literatures~\cite{droste2002analysis,doerr2012memory,Doerr2016}. The objective is to maximise the number of $1$s occurring in a binary string, i.e., for a given $n$-bit string $\mathbf{x}$, the fitness function to maximise for that string is given by
\begin{equation}
f(\x)=\sum_{i=1}^{n} x_i, \label{eqn:NoiseFreeFitnessFunction}
\end{equation}
where $x_i$ denotes the $i^{th}$ bit in the string $\mathbf{x}$, and is either 1 or 0.

\subsection{Noisy OneMax}\label{sec:nonemax}

In this work, we study a noisy variant of the 
OneMax problem, where the fitness function is corrupted by some additive unbiased normally distributed noise with constant variance, formalised in \eqref{eqn:noisyFitnessValue}.
\begin{equation}
f'(\x)=f(\x)+ \G(0,\sigma^2),\label{eqn:noisyFitnessValue}
\end{equation}
with $\G(0,\sigma^2)$ denoting a Gaussian noise, with mean $0$ and variance $\sigma^2$.
From now on, we will 
use $f'^{(i)}(\x)$ to denote the $i^{th}$ call to noisy fitness function on $\x$.
At each call, the noise is independently sampled from $\G(0,\sigma^2)$ and takes effect after the true fitness is evaluated.

There is a variety of noise models for the 
OneMax problem, as described in the next section. To avoid confusion, the term ``noisy OneMax'' refers to \eqref{eqn:noisyFitnessValue}, the noisy variant used in this paper. Also, the ``noisy'' refers to the noise in the fitness function's ability to read the true fitness of the genome; not the mutations which are deliberately applied to the genome, which could be interpreted as a second kind of ``noise''.

\subsection{Related work}\label{sec:rw}
Different noise models for the 
OneMax problem have been studied before.
\paragraph{Proportionate selection is noise invariant}
Our noise model is the same as the one used in~\cite{miller1996genetic}. Miller and Goldberg~\cite{miller1996genetic} used a GA with $\mu$ parents and $\lambda$ offsprings, 
and proved that increasing noise level did not affect selection pressure for Genetic Algorithms (GA) using \emph{fitness-proportionate selection}, in which each individual survives with a probability proportional to its fitness divided by the average fitness in the population~\cite{miller1996genetic,mitchell1998introduction}.

\paragraph{Resampling does not add benefit in low dimension}
Qian et al.~\cite{qian2014effectiveness} applied (1+1)-EA using mutation probability $p=\frac{1}{10}$ and at most $100$ resamplings to a $10$-bit noisy OneMax problem corrupted by additive Gaussian noise with variance $100$ (\eqref{eqn:noisyFitnessValue} with $\sigma^2=100$), and concluded that resampling wasn't beneficial for the (1+1)-EA 
in optimising the noisy OneMax problem.
In their model, resampling had the effect of lowering the variance of the noise to $1$, which is exactly as big as the lower bound of the differences between distinct noise-free fitness values. With that level of resampling and problem dimension, the problem is still difficult to solve.  We present, later in this paper, our application of RMHC using larger resampling numbers in noisy OneMax, with noise variance $\sigma^2=1$ and do observe that resampling adds a significant benefit when the dimension is above $10$ (Section \ref{sec:xp}).

\paragraph{Solving high dimension noisy OneMax}
Sastry et al.~\cite{sastry2007towards} designed a fully parallelised, highly-efficient compact Genetic Algorithm (cGA) to solve very high dimension problems, and compared it to RMHC on a noisy OneMax problem using noise variance depending linearly on the length of string, to allow for more difficult problems, for the reason that a randomly initiated $n$-bit string has fitness variance $\frac{n}{4}$. RMHC \emph{without} resampling performed poorly when the problem dimension is higher than $10,000$.

\paragraph{Noise takes effect before the evaluation}
Droste~\cite{droste2004analysis} defined a \emph{One-bit noise} model for the 
OneMax problem, in which during every fitness evaluation of the bit-string $\x$, there was exactly one uniformly-chosen random bit mis-read with probability $p'$. 
Hence the measured noisy fitness values are equivalent to replacing the Gaussian noise term in \eqref{eqn:noisyFitnessValue} by an appropriate discrete random variable taking values from $\{-1,0,1\}$.  
Under this scheme, Droste showed that (1+1)-EA with mutation probability $p=\frac1n$ could optimise a $n$-bit OneMax problem corrupted by \emph{One-bit noise}, with high probability, in polynomial time if $p'$ is $O(\frac{\log(n)}{n})$.

\section{Random Mutation Hill-Climbing in noisy context}\label{sec:rmhcr}

When RMHC is applied to the noisy OneMax fitness function \eqref{eqn:noisyFitnessValue}, a mutation of the $k^{th}$ gene refers to flipping the $k^{th}$ bit of the string. The standard deviation of the second term in \eqref{eqn:noisyFitnessValue} (the explicit noise term) is of the same order of magnitude as the noise in the first term of \eqref{eqn:noisyFitnessValue} (the OneMax fitness term) introduced by mutations from the RMHC algorithm. This extremely poor signal-to-noise ratio would cause major problems for hill-climbing strategies such as RMHC. Hence, 
we use RMHC with resampling, applied to the noisy OneMax problem
, so as to try to reduce the unwanted variance, and to allow the hill climber to work.

\subsection{Noise-free case}
Algorithm \ref{algo:noisefree} recalls the generic RMHC, without any resampling. This is suitable for noise-free problems.
\def\nouse{To solve a noise-free problem, no resampling is necessary. The variable $N$ counts the number of fitness evaluations made. Since line 6 makes two fitness evaluations (i.e. two calls to $f$), $N$ is incremented by 2 each generation. In some situations it is possible to modify the code to store the current best fitness value $f(\mathbf{x})$ instead of repeatedly reevaluating it. 

\begin{algorithm}[t]
\caption{\label{algo:noisefree}Random Mutation Hill-Climbing algorithm (RMHC).}
\begin{algorithmic}[1]
\Require{$n \in \N^*$: genome length (problem dimension)}
\Require{$\mathcal{X}$: search space}
\Require{$f: \mathcal{X} \mapsto \R$: fitness function}
\State{Randomly initialise a genome $\mathbf{x} \in \mathcal{X}$}
\State{$N \gets 0$}	\Comment{Total evaluation count so far}
\While{time not elapsed}
    \State{Uniformly randomly select $k \in \{1,\dots,n\}$} 
    \State{$\mathbf{y} \gets$ new genome by mutating the $k^{th}$ gene of $\mathbf{x}$}
    \If{$f(\mathbf{y}) \geq f(\mathbf{x})$}
    	\State{$\mathbf{x} \gets \mathbf{y}$}	\Comment{Update the best-so-far genome}
    \EndIf
    \State{$N \gets N + 2$}\Comment{Update evaluation count}
\EndWhile
\State{\Return{$\mathbf{x}$}}
\end{algorithmic}
\end{algorithm}
}

Here we have assumed the fitness value of the best-so-far genome could be stored after each cycle. If, alternatively, no space was allocated to store that fitness value, or if fitness evaluations can only be made by directly comparing two individuals, then the best-so-far genome's fitness would need to be re-evaluated at each generation, thus raising the algorithm's ``evaluation count'', $N$, by a factor of approximately 2.
\begin{algorithm}[t]
\caption{\label{algo:noisefree}Random Mutation Hill-Climbing algorithm (RMHC). No resampling is performed}.
\begin{algorithmic}[1]
\Require{$n \in \N^*$: genome length (problem dimension)}
\Require{$\mathcal{X}$: search space}
\Require{$f: \mathcal{X} \mapsto \R$: fitness function}
\State{Randomly initialise a genome $\mathbf{x} \in \mathcal{X}$}
\State{$bestFitSoFar \gets f(\mathbf{x})$}
\State{$N \gets 1$}	\Comment{Total evaluation count so far}
\While{time not elapsed}
    \State{Uniformly randomly select $k \in \{1,\dots,n\}$} 
    \State{$\mathbf{y} \gets$ new genome by mutating the $k^{th}$ gene of $\mathbf{x}$}
    \State{$Fit_{\mathbf{y}} \gets f(\mathbf{y})$}
    \State{$N \gets N + 1$}\Comment{Update evaluation count}
    \If{$Fit_{\mathbf{y}} \geq bestFitSoFar$}
    	\State{$\mathbf{x} \gets \mathbf{y}$}	\Comment{Update the best-so-far genome}
        \State{$bestFitSoFar \gets Fit_{\mathbf{y}}$}
    \EndIf
\EndWhile
\State{\Return{$\mathbf{x}$}}
\end{algorithmic}
\end{algorithm}

\subsection{Noisy case}\label{sec:noisy}
The previous RMHC algorithm (Algorithm \ref{algo:noisefree}) was applicable to deterministic fitness functions.  
Algorithm \ref{algo:noisy} extends this RMHC algorithm to be applicable to noisy fitness functions. It achieves this extension by using \emph{resampling}, so that each genome is evaluated multiple times, so as to reduce the effect of noise which might interfere with hill climbing.

Additionally, if the statistics of the best-so-far genome can be stored, instead of comparing directly the fitness values of the offspring to the fitness of the parent (the best-so-far genome), the average fitness value of the best-so-far genome in the history is compared at each generation (line 11 of Algorithm \ref{algo:noisy}).

\begin{algorithm}[t]
\caption{\label{algo:noisy}RMHC modified to include resampling, suitable for the noisy case. $f'^{(i)}(\mathbf{x})$ denotes the $i^{th}$ call to the noisy fitness function on search point $\mathbf{x}$.}
\begin{algorithmic}[1]
\Require{$n \in \N^*$: genome length (problem dimension)}
\Require{$\mathcal{X}$: search space}
\Require{$f: \mathcal{X} \mapsto \R$: fitness function}
\Require{$r \in \N^*$: Resampling number}
\State{Randomly initialise a genome $\mathbf{x} \in \mathcal{X}$}
\State{$bestFitSoFar \gets 0$}
\State{$M \gets 0$}	\Comment{Evaluation count for the latest best-so-far genome}
\State{$N \gets 0$}	\Comment{Total evaluation count so far}
\While{time not elapsed}
    \State{Uniformly randomly select $k \in \{1,\dots,n\}$}
 \State{$\mathbf{y} \gets$ new genome by mutating the $k^{th}$ gene of $\mathbf{x}$}
 \State{$Fit_{\mathbf{x}} \gets \frac{1}{r}\sum _{i=1}^{r} f'^{(i)}(\mathbf{x})$}
    \State{$Fit_{\mathbf{y}} \gets \frac{1}{r}\sum _{i=1}^{r} f'^{(i)}(\mathbf{y})$}
    \State{$N \gets N + 2r$}	\Comment{Update evaluation count}
    \State{$averageFitness_{\mathbf{x}} \gets \frac{bestFitSoFar\times M+Fit_{\mathbf{x}}\times r}{M+r}$} 
    \If{$Fit_{\mathbf{y}} \geq averageFitness_{\mathbf{x}}$}
    	\State{$\mathbf{x} \gets \mathbf{y}$}	\Comment{Update the best-so-far genome}
   		\State{$bestFitSoFar \gets Fit_{\mathbf{y}}$} 
        \State{$M \gets r$}
    \Else
    	\State{$bestFitSoFar \gets averageFitness_{\mathbf{x}}$} 
        \State{$M \gets M + r$}
    \EndIf
\EndWhile
\State{\Return{$\mathbf{x}$}}
\end{algorithmic}
\end{algorithm}

\section{Analysis of optimal resampling number in noisy OneMax problem}\label{sec:analysis}

This section analyses the application of the modified RMHC algorithm (Algorithm \ref{algo:noisy}) to the noisy OneMax problem, with fixed noise level, by describing it as a Markov process. Several variants of Algorithm \ref{algo:noisy} are considered in Section \ref{sec:benefit}, including analysis of varying the resampling number, and of the effect of storing a statistic of the best-so-far genome versus not storing it. The benefit of storing the statistics of the best-so-far genome is illustrated in Section \ref{sec:benefit}. Section \ref{sec:optimalr} derives an expression for the length of Markov chain that represents the learning algorithm, and the optimal level of resampling is calculated and displayed, in Section \ref{sec:theoreticalResults}.

We restrict the analysis in this section to a fixed noise level $\sigma^2=1$.  This noise level is a significant challenge for RMHC, as the difference between the parent's fitness and the one of the offspring is always $1$ in our model. This motivates resampling, since resampling a candidate solution $r$ times can reduce the variance of the noise by a factor of $\frac1r$.  The extension of this analysis to the general $\sigma$ would be straightforward.


\subsection{Markov chain description for noisy OneMax}\label{sec:markovChainDescription}

We consider a $n$-bit OneMax problem with constant variance Gaussian noise, $\sigma^2=1$.   Throughout this section, we summarise the n-bit OneMax state vector by a single scalar number, $i$, equal to the number of ones in the n-bit string. As the RMHC algorithm makes mutations to the full bit string, the compressed state-representation, $i$, will change by $\pm 1$.  The transition probabilities for the change in $i$ are dependent only on the scalar $i$.  Hence the evolution of the variable $i$ is modeled by a Markov process.

After each mutation is initially made by the RMHC algorithm, the fitness of that mutated bit string is evaluated using \eqref{eqn:noisyFitnessValue}, and the RMHC algorithm will either accept or reject that mutation.

Let $p_{TA}$, $p_{FA}$, $p_{TR}$ and $p_{FR}$ denote the probability of true acceptance, false acceptance, true rejection and false rejection, respectively, for the RMHC algorithm to accept or reject any given mutation.  These four probabilities depend on the resampling strategy employed by the RMHC algorithm, and are derived in Section \ref{sec:acceptanceProabilities}.  However, since complementary probability pairs must sum to one, we do generally have that,
\begin{eqnarray}
p_{FA} &=& 1-p_{TA},\label{eq:fa}\\
p_{FR} &=& 1-p_{TR}.\label{eq:fr}
\end{eqnarray}

Assuming these acceptance and rejection probabilities are known, we can then derive the Markov state transition probabilities as follows:

For any state scalar $i \in \{0,1,\dots,n-1\}$, the corresponding OneMax bit string has $i$ ones and $(n-i)$ zeros. Therefore the probability of uniformly randomly choosing a zero bit is $\frac{n-i}{n}$. Hence, for RMHC to make an improvement to the genome, it must randomly choose one of these zero bits and that mutation must be accepted. Therefore the transition probability from state $i$ to state $i+1$ in one generation is:
\begin{equation}
\P[S_{t+1}=i+1 | S_{t}=i] = \frac{n-i}{n}p_{TA}.\label{eqn:probabilityTransitionImprovement}
\end{equation}

Similarly, for RMHC to make the genome worse, it must choose a one bit (with probability $\frac{i}{n}$) and flip it to a zero, and that mutation must be accepted, with probability $p_{FA}$. Hence we obtain

\begin{equation}
\P[S_{t+1}=i-1 | S_{t}=i] = \frac{i}{n}p_{FA}.
\end{equation}

For an 
RMHC mutation to make no progress in the genome, the mutation must be rejected. This could mean a one bit is chosen (with probability $\frac{i}{n}$) and rejected (with probability $p_{TR}$), or it could be that a zero bit is chosen (with probability $\frac{n-i}{n}$) and rejected (with probability $p_{FR}$). Hence we obtain
\begin{equation}
\P[S_{t+1}=i | S_{t}=i] = \frac{i}{n}p_{TR}+\frac{n-i}{n}p_{FR}.\label{eqn:probabilityTransitionNoChange}
\end{equation}

\begin{figure*}[h]
\centering
\includegraphics[width=.8\linewidth]{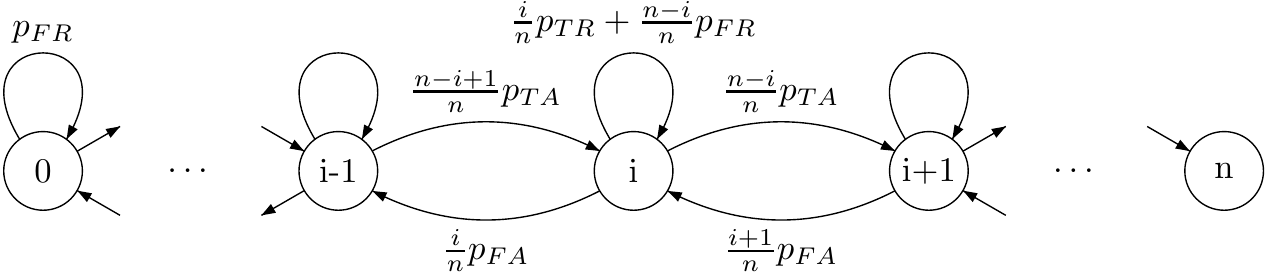}
\caption{\label{fig:state}Markov states and transitions in a noisy OneMax problem
. Markov state 
``$i$'' refers to a OneMax bit-string with $i$ ones and $n-i$ zeros in the string.}
\end{figure*}

The probabilities given by \eqref{eqn:probabilityTransitionImprovement}-\eqref{eqn:probabilityTransitionNoChange} appear on the three arrows emanating from the central node ``$i$'' in the Markov chain shown in \figurename \ref{fig:state}.  The Markov chain's absorption state is state $n$, since the OneMax problem is solved and terminates as soon as $i=n$ is reached.

\def\nouse{
{\color{red} Together with Equations \ref{eq:tr}, \ref{eq:fa} and \ref{eq:fr}, $\forall i \in \{1,2,\dots,n-1\}$ the transition probability matrix of two adjacent states is:
\[ B= \left[ \begin{array}{cc}
\frac{i}{n}p_{TA} + \frac{n-i}{n}(1-p_{TA}) & \frac{n-i}{n}p_{TA} \\
\frac{i+1}{n}(1-p_{TA}) & \frac{i+1}{n}p_{TA} + \frac{n-i-1}{n}(1-p_{TA}) \end{array} \right].\] Let's delete this red paragraph - it's not justified why it's 2 by 2 matrix, nor where the elements come from, and it's not used in any analysis AFAIK}}

\subsection{Rejection and acceptance probabilities for noisy OneMax problem with RMHC}\label{sec:benefit} \label{sec:acceptanceProabilities}
The Markov Process described in the previous subsection relied upon knowledge of the acceptance and rejection probabilities $p_{TR}$ and $p_{TA}$, which are dependent on the RMHC resampling method chosen.


We discuss three RMHC resampling cases here, and calculate the corresponding acceptance/rejection probabilities.  We consider three separate cases:

\subsubsection{No resampling, no statistic of the best-so-far genome is stored}\label{sec:case1}
If no statistic of the best-so-far genome is stored, the best-so-far one needs to be re-evaluated once at each generation in the case without resampling.

First we assume that the newly generated genome $\mathbf{y}$ is better than the current genome $\mathbf{x}$, i.e. $f(\mathbf{y}) > f(\mathbf{x})$, where this fitness function is the noise-free version given by \eqref{eqn:NoiseFreeFitnessFunction}. Since $f(\mathbf{x})$ is the true fitness of string $\mathbf{x}$, we have $f(\mathbf{x})=i$, the number of ones in string $\mathbf{x}$. Since the two genomes $\mathbf{x}$ and $\mathbf{y}$ are evaluated using the noisy fitness function \eqref{eqn:noisyFitnessValue}, and since we are comparing $\mathbf{x}$ and $\mathbf{y}$ without resampling or storing the statistics of the best-so-far genome, the probability of true acceptance is 
\begin{eqnarray*}
p_{TA} &=& \P(f'(\mathbf{y}) > f'(\mathbf{x}) | f(\mathbf{y})=i+1, f(\mathbf{x})=i)\\
&=& \P(\omega_y + 1 > \omega_x)=\P(\omega_y - \omega_x > -1).
\end{eqnarray*}
where $\omega_y$ and $\omega_x$ are 
independent samples from $\G(0,1)$, thus $\omega_y - \omega_x$ $\sim \G(0,2)$. Then,
\begin{eqnarray}
p_{TA} &=& \P(\omega_y - \omega_x > -1) = \P(\omega_y - \omega_x \leq 1)\nonumber \\
&=& CDF_{Gaussian(0,2)}(1) = \frac12 + \frac12 erf(\frac12). \label{eqn:ptaCase1}
\end{eqnarray}
Respectively, if the newly generated genome $\mathbf{y}$ is worse than the current genome $\mathbf{x}$, i.e. $f(\mathbf{y}) < f(\mathbf{x})$, when comparing two genomes without resampling or storing the statistics of the best-so-far genome, the probability of true rejection is 
\begin{eqnarray}\label{eq:tatr}
p_{TR} &=& \P(f'(\mathbf{y}) < f'(\mathbf{x}) | f(\mathbf{y})=i-1, f(\mathbf{x})=i)\nonumber\\
&=& \P(\omega_y - 1 < \omega_x)= \P(\omega_y - \omega_x < 1) = p_{TA}.
\end{eqnarray}

Therefore, for this situation of RMHC with no resampling and no storage of the best-so-far fitness, we can for example find the probability of transferring from state $i$ to state $i+1$, as follows:
\begin{align*}
\P[S_{t+1}=i+1 | S_{t}=i] &= \frac{n-i}{n}p_{TA} &\text{(by \eqref{eqn:probabilityTransitionImprovement})}\\
&=\frac{n-i}{n}\left(\frac12 + \frac12 erf(\frac12)\right)&\text{(by \eqref{eqn:ptaCase1}).}
\end{align*}

\subsubsection{When comparing two genomes using $r$ resamplings without storing the statistics of the best-so-far genome}\label{sec:case2}
If each genome can be re-evaluated $r>1$ times, i.e., $r$ resamplings are used, but still no statistic of the best-so-far genome is stored, $Fit_{\mathbf{x}}$ and $Fit_{\mathbf{y}}$ are compared at each generation (lines 8 and 9 of Algorithm \ref{algo:noisy}). Therefore, the variance of $Fit_{\mathbf{x}}$ and $Fit_{\mathbf{y}}$, given $\mathbf{x}$ and $\mathbf{y}$, are $\frac{1}{r}$. Then,
\begin{eqnarray*}
p_{TA} &=& \P(Fit_{\mathbf{y}} > Fit_{\mathbf{x}} | f(\mathbf{y})=i+1, f(\mathbf{x})=i)\\
&=& \P(\omega'_y + 1 > \omega'_x)=\P(\omega'_y - \omega'_x > -1).
\end{eqnarray*}
where $\omega'_y$ and $\omega'_x$ are 
independent samples from $\G(0,\frac{1}{r})$, thus $\omega_y - \omega_x$ $\sim \G(0,\frac{2}{r})$. Then,
\begin{eqnarray}
p_{TA} &=& \P(\omega'_y - \omega'_x > -1)=\P(\omega'_y - \omega'_x \leq 1)\nonumber \\
&=& CDF_{Gaussian(0,\frac{2}{r})}(1) = \frac12 + \frac12 erf(\frac{\sqrt{r}}{2}). \label{eqn:ptaCase2}
\end{eqnarray}
Similarly, $p_{TR}=p_{TA}$ holds in this case.

The probability of transferring from state $i$ to state $i+1$ in one generation is $\frac{n-i}{n}(\frac12 + \frac12 erf(\frac{\sqrt{r}}{2}))$, larger than the probability 
in the previous 
case without resampling (Section \ref{sec:case1}).

Therefore, the probability of true acceptance is improved by resampling genomes.

\subsubsection{When comparing two genomes using $r>1$ resamplings and the statistics of the best-so-far genome}\label{sec:case3}
Additionally, if one has access to $averageFitness$, the average fitness value of the best-so-far genome, and $M$, the number of times that the best-so-far genome has been evaluated, by line 11 of Algorithm \ref{algo:noisy}
\begin{equation}
averageFitness_{\mathbf{x}} \gets \frac{bestFitSoFar\times M+Fit_{\mathbf{x}}\times r}{M+r},
\end{equation}
with $r>1$. Therefore, the variance of $averageFitness_{\mathbf{x}}$, given $\mathbf{x}$, is $\frac{1}{M+r}$. The variance of $Fit_{\mathbf{y}}$, given $\mathbf{y}$, is $\frac{1}{r}$.

The probability of true acceptance is
\begin{align*}
p_{TA} &= \P( Fit_{\mathbf{y}} > averageFitness_{\mathbf{x}} | f(y)=i+1, f(x)=i)\\
&= \P(\omega'_y + 1 > \omega''_x)= \P(\omega'_y - \omega''_x > -1).
\end{align*}
where $\omega'_y$ and $\omega''_x$ are independent samples from $\G(0,\frac{1}{r})$ and $\G(0,\frac{1}{M+r})$, respectively. Thus, $\omega'_y - \omega''_x$ $\sim \G(0, \frac{M+2r}{r(M+r)})$. Then,
\begin{eqnarray}\label{eq:pta3}
p_{TA} &=& \P(\omega'_y - \omega''_x > -1)\nonumber \\
&=& \P(\omega'_y - \omega''_x \leq 1)\nonumber \\
&=& CDF_{Gaussian(0,\frac{M+2r}{r(M+r)})}(1)\nonumber \\
&=& \frac12 + \frac12 erf(\sqrt{\frac{r(M+r)}{2(M+2r)}})
\end{eqnarray}
Similarly, $p_{TR}=p_{TA}$ holds in this case.

The probability of transferring from state $i$ to state $i+1$ in one generation is $\frac{n-i}{n}(\frac12 + \frac12 erf(\sqrt{\frac{r(M+r)}{2(M+2r)}}))$ with $M \geq r >1$, larger than the probabilities 
in the previous cases (Sections \ref{sec:case1} and \ref{sec:case2}).

Therefore, the probability of true acceptance is improved by resampling genomes and storing the statistics of the best-so-far genome.
However, 
the trade-off between the total evaluations and the accuracy needs to be considered.

\subsection{Markov chain analysis}\label{sec:optimalr}

Now that we have described the Markov chain in Section \ref{sec:markovChainDescription}, and derived the acceptance and rejection probabilities for the noisy OneMax problem in Section \ref{sec:acceptanceProabilities}, we next derive an analytical expectation for the full trajectory length for solving the noisy 
OneMax problem, using RMHC with resamplings, starting from a bit string full of zeros. To simplify analysis, no stored statistic is considered ($M=0$), thus we consider the second case discussed previously in Section \ref{sec:case2}.

The Markov chain length can be found by defining the notation $\et{m}{l}$ to mean the expectation of the number of generations required to get from a Markov state with value $i=l$ to a Markov state with value $i=m$.

By considering the three arrows that emanate from the central $i$ node in \figurename \ref{fig:state}, we can form an algebraic expression for $\et{i+1}{i}$, as follows:
\begin{align} 
\et{i+1}{i}=& \frac{n-i}{n}p_{TA}\nonumber \\
&+ \left(\frac{i}{n}p_{TR} + \frac{n-i}{n}p_{FR}\right)\left(1+\et{i+1}{i}\right)\nonumber\\
&+ \frac{i}{n}p_{FA}\left(1+\et{i+1}{i-1}\right). \label{eqn:markovRecursion1}
\end{align}

The three terms in \eqref{eqn:markovRecursion1} correspond to the three arrows from node $i$ in \figurename \ref{fig:state}, pointing to nodes $i+1$, $i$ and $i-1$, respectively.

Since it is always necessary to go from state $i-1$ to $i+1$ via their middle state $i$, we can form the relationship
\begin{align*}
\et{i+1}{i-1} &\equiv \et{i}{i-1}+\et{i+1}{i}.
\end{align*}
Furthermore, reusing \eqref{eq:fa} and \eqref{eq:fr}, i.e. $p_{FA}=(1-p_{TA})$, $p_{FR}=(1-p_{TR})$, together with $p_{TR}=p_{TA}$ from Section \ref{sec:acceptanceProabilities}, we get:
\begin{eqnarray}\label{eq:expN}
&&\et{i+1}{i}= \frac{n-i}{n}p_{TA}\nonumber \\
&+& \left(\frac{i}{n}p_{TA} + \frac{n-i}{n}(1-p_{TA})\right)\left(1+\et{i+1}{i}\right)\nonumber\\
&+& \frac{i}{n}(1-p_{TA})\left(1+\et{i}{i-1}+\et{i+1}{i}\right).
\end{eqnarray}

Solving \ref{eq:expN} for $\et{i+1}{i}$, which appears three times in that equation, we get:
\begin{equation}
\et{i+1}{i} = \frac{i(1-p_{TA})}{(n-i)p_{TA}} \et{i}{i-1}+ \frac{n}{(n-i)p_{TA}}. \label{eqn:markovRecursion3}
\end{equation}

This is a recursive equation that defines the $i^{th}$ term in terms of the $(i-1)^{th}$ term.  To terminate this recursion, an explicit expression for $\et{1}{0}$ can be found by considering the number of generations required to get from $i=0$ (i.e. a string with all zeros in it) to $i=1$ (i.e. a string with exactly one 1 in it). This is given by
\begin{eqnarray}\label{eq:zeros}
\et{1}{0}=\frac{1}{p_{TA}}.
\end{eqnarray}

\eqref{eqn:markovRecursion3} and \eqref{eq:zeros} form a recursion that easily can be unrolled computationally, and it gives us the exact theoretical number of transitions to traverse from one Markov state $i$ to the adjacent Markov state $i+1$. Therefore to calculate the total number of generations required to solve the noisy OneMax problem from an initial Markov state $i=0$, we need to calculate $\et{n}{0}$. This can be expanded by adding all of the intermediate state transitions to get
\begin{equation}
\et{n}{0}=\et{1}{0} + \cdots + \et{n}{n-1} \label{eqn:totalMarkovTrajectoryLength}
\end{equation}

\eqref{eqn:totalMarkovTrajectoryLength} completes the theoretical analysis of the number of steps required by RMHC to solve the noisy 
OneMax problem, from an initial bit string of zeros. Note that each term of this sum needs a solution to the recursive equations given by \eqref{eqn:markovRecursion3} and \eqref{eq:zeros}, but with careful caching, the whole sum can be evaluated in $O(n)$ steps.

It's notable that even though the above Markov chain analysis was aimed at Case 2 
(Section \ref{sec:case2},  i.e. with resampling and no statistic stored), it also holds when there is no resampling, i.e., $r=1$ (the first case detailed in Section \ref{sec:case1}). In both cases, $p_{TA}$ is deterministic given the resampling number $r$. However, Case 3 (Section \ref{sec:case3}) is not as straightforward to analyse, because in that case that the average fitness value ($averageFitness_{\mathbf{x}}$) depends on the evaluation number ($M$), and $M$ is stochastic. Hence, $p_{TA}=\frac12 + \frac12 erf(\sqrt{\frac{r(M+r)}{2(M+2r)}})$ \eqref{eq:pta3} 
changes at each generation of RMHC. 
Case 3 has not been analysed in this paper, but it 
would be interesting to do so in the future.
 
\subsection{Analytical results of RMHC in noisy OneMax problem} \label{sec:theoreticalResults}

\begin{figure}[h]
\centering
\includegraphics[width=\linewidth]{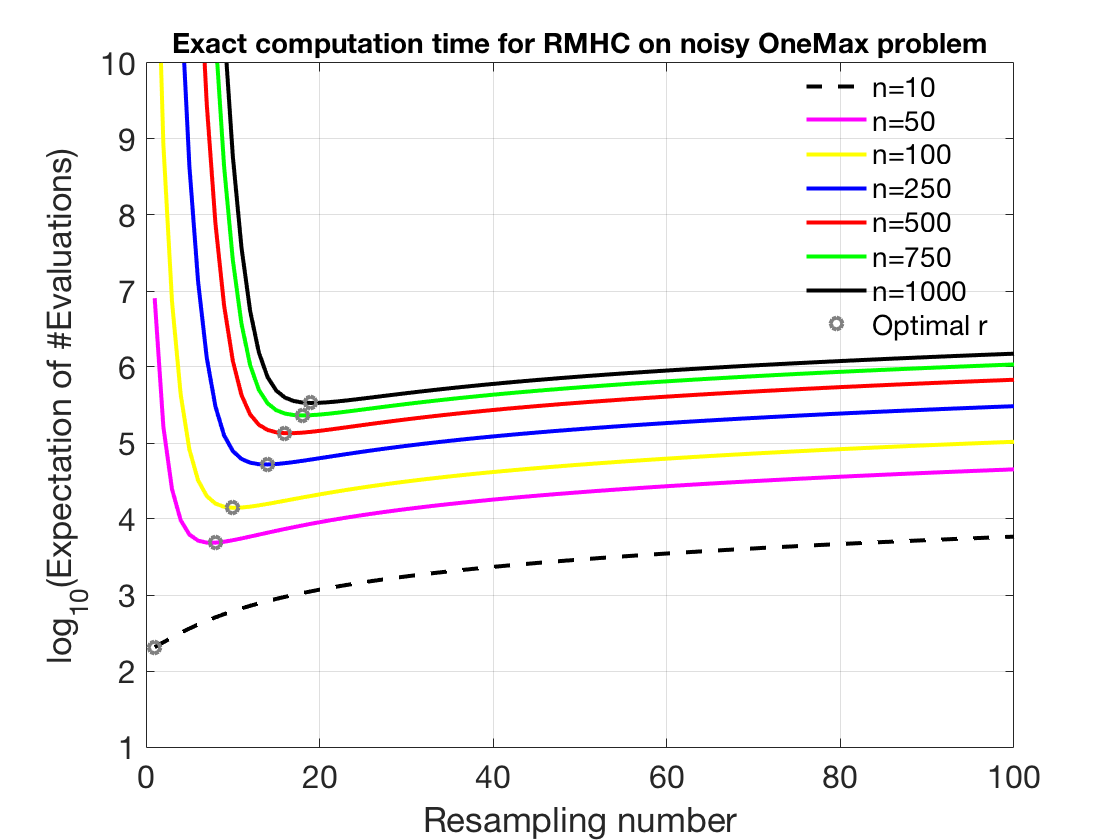}
\caption{\label{fig:expE}Exact expectation of the number of fitness evaluations required by RMHC to solve the noisy 
OneMax problem, using different resampling numbers. The initial OneMax strings were all zeros, and the RMHC algorithm did not store the fitness of the best-so-far genome (case 2 described in Section \ref{sec:case2}). These curves were computed using \eqref{eqn:markovRecursion3}-\eqref{eq:sum}, and $p_{TA}$ defined by \eqref{eqn:ptaCase2}.
The grey circle on each curve indicates the optimal resampling number, which increases with the problem dimension.}
\end{figure}

At each generation of the actual RMHC algorithm, $2r$ fitness-function evaluations are required, so that the total number of fitness evaluations required to solve the noisy 
OneMax problem is 
\begin{equation}\label{eq:sum}
2r\et{n}{0}.
\end{equation}
This result is shown graphically in \figurename \ref{fig:expE}, with various number of resamplings ($r$) and problem dimension ($n$), under the assumption that no statistic is stored for the fitness of the best-so-far genome (case 2 described in Section \ref{sec:case2}).

\begin{figure}[h]
\centering
\includegraphics[width=\linewidth]{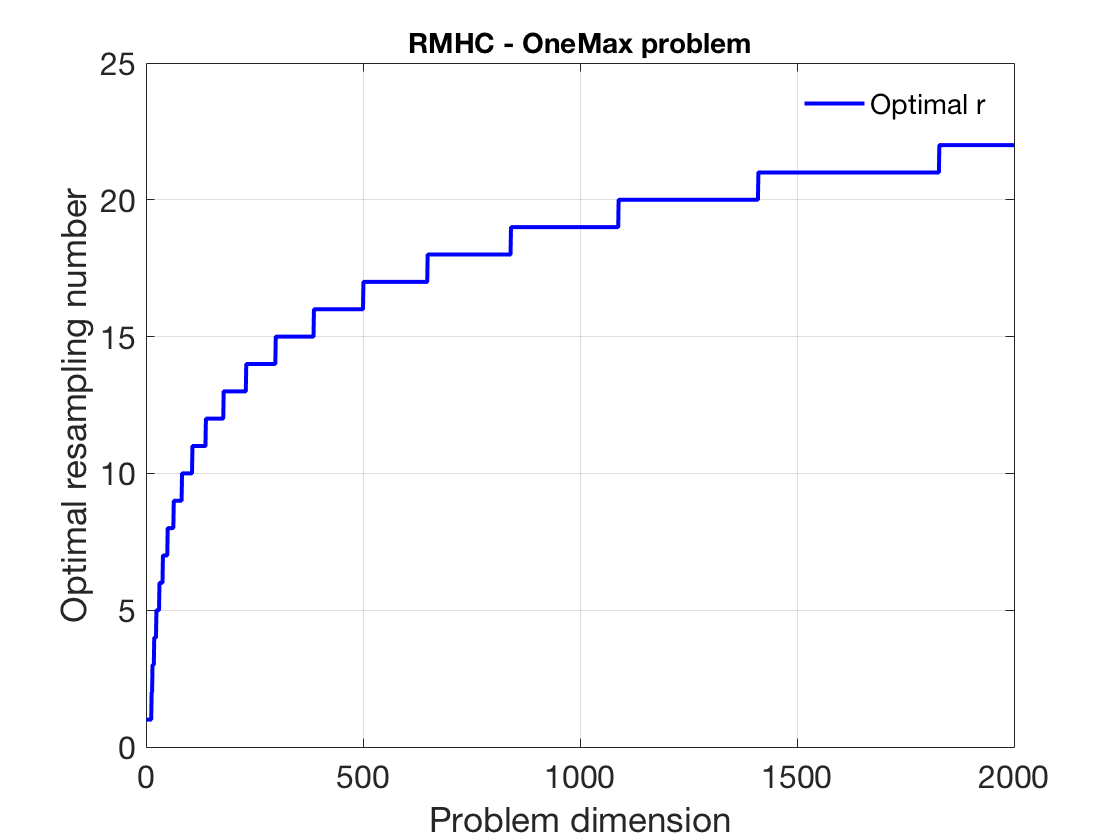}
\caption{\label{fig:op}Exact theoretical optimal resampling number required by RMHC, to solve the noisy OneMax problem, without storing a statistic of the fitness of the best-so-far genome (case 2 described in Section \ref{sec:case2}).}
\end{figure}

As can be seen from the location of the minima in \figurename \ref{fig:expE}, indicated by the small grey circles, the optimal resampling number increases with the problem dimension. The exact optimal number of resamplings to make in different dimensions is displayed in \figurename \ref{fig:op}.


\section{Experimental results}\label{sec:xp}

\begin{figure}
\centering
\begin{subfigure}[b]{\linewidth}
\includegraphics[width=\textwidth]{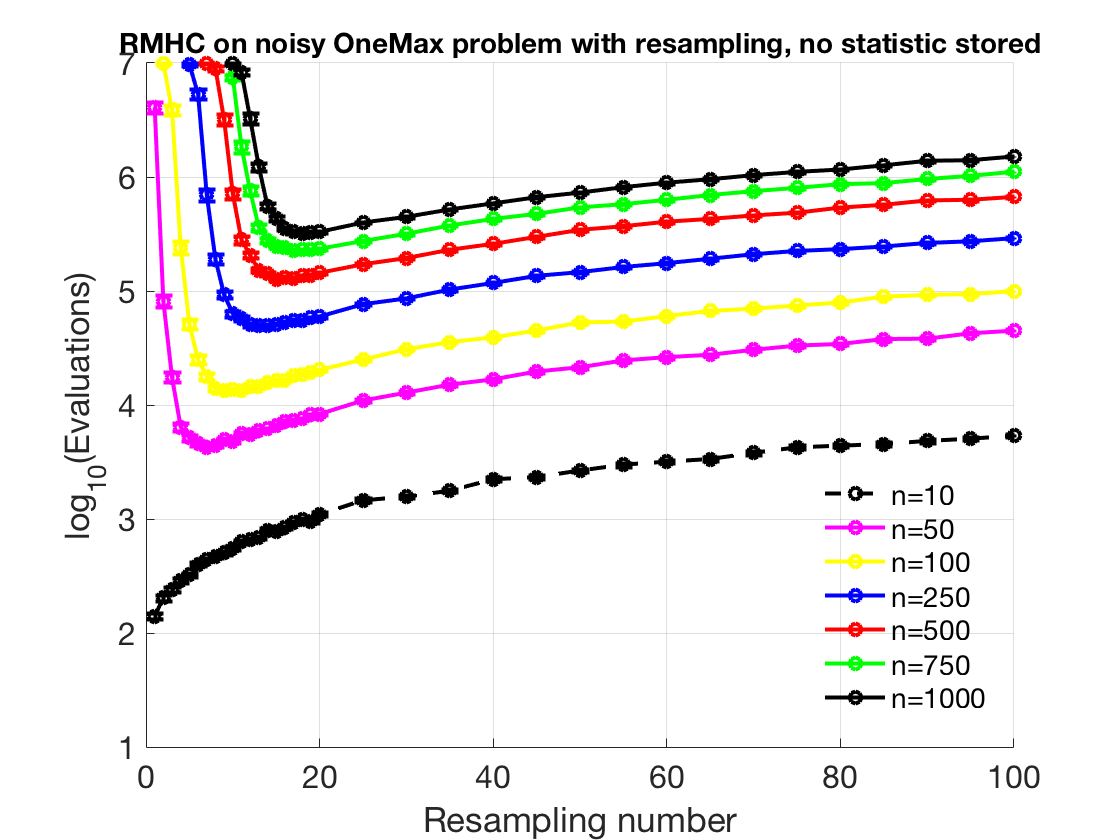}
\caption{\label{fig:rfrmhc2}Empirical number of evaluations consumed by RMHC in the noisy OneMax problem, across different resampling numbers, without storing a statistic of the best-so-far genome. This empirically-obtained data is equivalent to the theoretical 
results shown in \figurename \ref{fig:expE}.}
\end{subfigure}
\begin{subfigure}[b]{\linewidth}
\includegraphics[width=\textwidth]{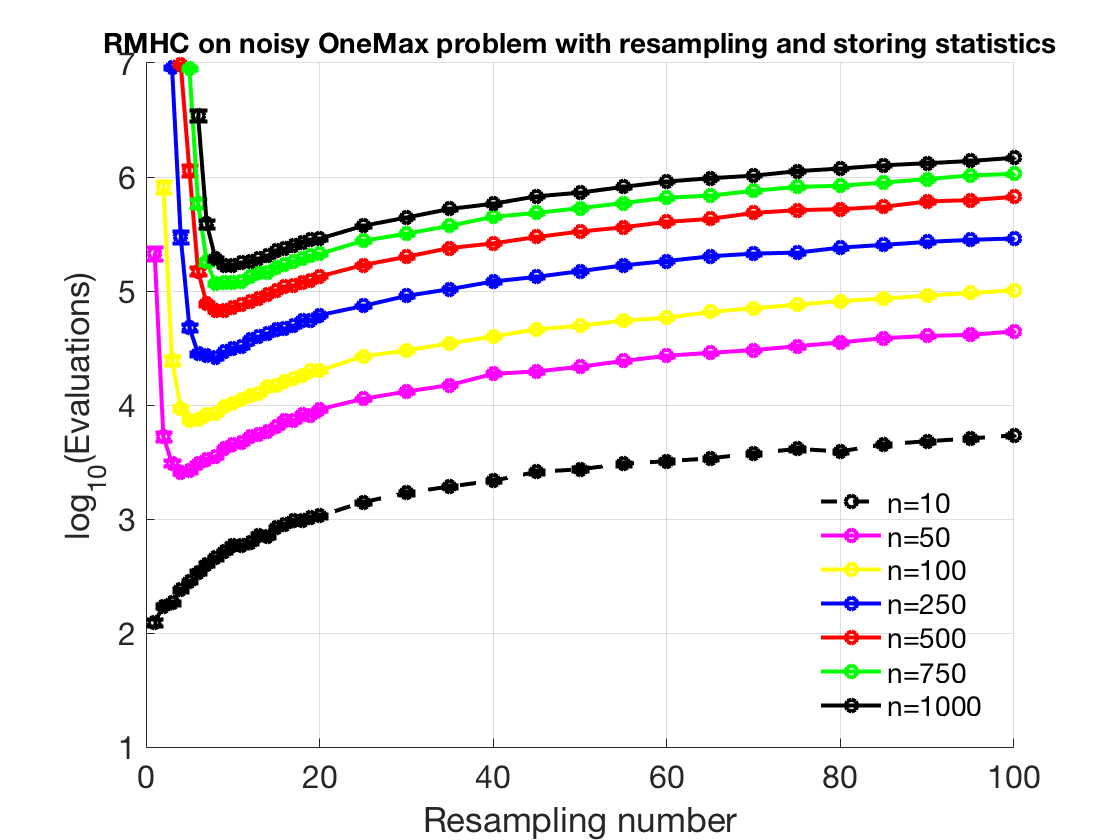}
\caption{\label{fig:rfrmhc3}Empirical number of evaluations consumed by RMHC in the noisy OneMax problem, across different resampling numbers, using the \emph{stored statistic} method. Fewer 
fitness evaluations are consumed compared to the case without stored statistic (shown in the above figure). Here the $n=1000$ curve is truncated on the left, showing missing data points, because in those cases an optimum was not reached within $10^7$ fitness evaluations, when using low resampling numbers ($r=1,2,3,4,5$).}
\end{subfigure}
\caption{\label{fig:rf}Performance of RMHC in the noisy OneMax problem with different numbers of resamplings. The maximal budget is $10^7$ fitness evaluations.
Note that the y-axis is the 
logarithm of the evaluation number. Each experiment was repeated $100$ times, initialised by an all-zeros string. The standard error is too tiny to be seen.
In the noisy case, the optimal resampling number increases with the problem dimension.}
\end{figure}

To supplement the theoretical results of the previous and to prove the benefit of storing statistics of the best-so-far genome at each generation, we apply first RMHC on the OneMax problem in a noise-free case (\figurename \ref{fig:rmhc}), then evaluate the performance of RMHC on the OneMax problem with the presence of constant variance Gaussian noise (\figurename \ref{fig:rf}).
Each experiment was repeated $100$ times, initialised by an all zeros string.
The following may be observed from the figures:
\begin{figure}[h]
\centering
\includegraphics[width=\linewidth]{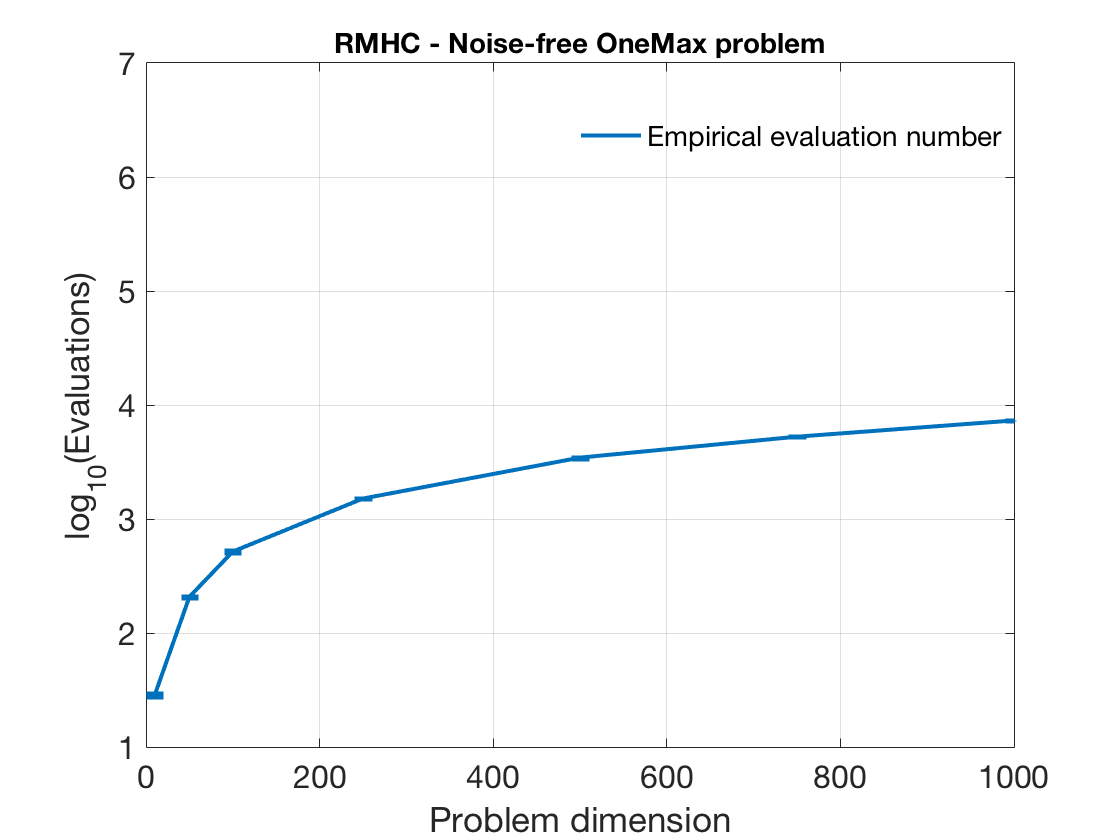}
\caption{\label{fig:rmhc}Empirical number of evaluations consumed by RMHC in the noise-free OneMax problem. No resampling is required in the noise-free case. Each experiment is repeated $100$ times initialised by an all zeros string. The standard error is too tiny to be seen.}
\end{figure}

\begin{itemize}
\item due to the noise, far more fitness evaluations are required to find the optimum in the noisy case than in the noise-free context;
\item the 
higher dimension problem required more fitness evaluations to reach the optimum;
\item in dimensions $\leq$ 10, no resampling leads to a reduction in the number of required fitness evaluations;
\item in higher dimensions ($n>10$), the optimal amount of resampling required increases with the problem dimension;
\item in higher dimensions ($n\gg10$), the evaluation number is significantly reduced by using the stored statistic.
\end{itemize}

Additionally, \figurename \ref{fig:rf} clearly shows the benefit of storing the average fitness value and the current evaluation number of the best-so-far genome at the end of each generation (case 3 described in Section \ref{sec:case3}).

\subsection{Validation of Theoretical results} \label{sec:validationTheoretical}

To validate the theoretical results and equations of Section \ref{sec:analysis}, Table \ref{tab} demonstrates a very close match between the theoretically obtained number of evaluations (derived from \eqref{eqn:markovRecursion3}-\eqref{eq:sum}) and the equivalent empirically calculated numbers (i.e. those found by actually running the RMHC algorithm repeatedly and averaging). This table hopefully validates the accuracy of our theoretical derivations 
and our numerical implementation.
\begin{table}
\centering
\caption{\label{tab}The expectation of fitness evaluation number required to reach the optimum (\eqref{eqn:markovRecursion3} and \eqref{eq:zeros}) in the noisy OneMax problem on dimension $10$ and the empirical average fitness evaluations consumed in $10,000$ trials.}
\begin{tabular}{ccc}
\hline
$r$ & Expected \#evaluations & Empirical \#evaluations\\
\hline
1 &205.8283 & 205.1998\\
2 &238.5264 & 239.7504\\
3 &276.3340 & 274.9920\\
4 &317.9576 & 317.8848\\
5 &362.4065 & 363.2520\\
10 &612.2250 & 611.0060\\
\hline
\end{tabular}
\end{table}

A further confirmation is shown by the equivalence of the curves shown in \figurename \ref{fig:rfrmhc2} to those shown in \figurename \ref{fig:expE}.

\section{Conclusion}\label{sec:conc}
This paper presents a noisy OneMax problem with additive constant variance Gaussian noise and analyses the optimal resampling number required by the Random Mutation Hill Climber (RMHC) to handle the noise.

The number of fitness evaluations required by RMHC to find the optimal solution in the noisy OneMax problem varies with the resampling number. In a very low-dimensional noisy OneMax problem (dimension 10, hence the string only has $1024$ possible values), the optimal value may be found by a random walk, and resampling can be counterproductive in these cases (it leads to the remarkable growth of the number of evaluations required to find the optimum as shown in \figurename \ref{fig:rf}).
However, in higher dimensions, resampling to reduce the noise is of critical importance, and makes the difference between success and failure. The optimal level of resampling increases with the dimension in the search space (as shown empirically in \figurename \ref{fig:expE}, and analytically in \figurename \ref{fig:op}). This is an interesting result, which for this particular benchmark problem, is in conflict with the observation by Qian et al. in \cite{qian2014effectiveness} that resampling was not beneficial.

RMHC is simple but efficient. The success of ($\mu$-$\lambda$)-GA, of which (1+1)-EA can be seen as a variant, depends on one or more parameters, such as the size of the population, $\lambda$, the number of parents, $\mu$, and the crossover and mutation operators. RMHC does not have such details to adjust and is therefore simpler to apply.
Due to its efficiency and simplicity RMHC should be considered as a
useful tool for expensive optimisation tasks.




%
\balance
\bibliographystyle{IEEEtran}
\bibliography{banditRMHC_TEVC}
\balance
\end{document}